%% file: acl_latex.tex
\documentclass[11pt]{article}

\usepackage[preprint]{acl}

\usepackage{times}
\usepackage{latexsym}

\usepackage[T1]{fontenc}

\usepackage[utf8]{inputenc}

\usepackage{microtype}

\usepackage{inconsolata}

\usepackage{graphicx}
\usepackage{amsmath}
\usepackage{amssymb}
\usepackage{algorithm}
\usepackage{algpseudocode}
\usepackage{booktabs}
\usepackage{multirow}

\usepackage{enumitem}

\title{DA-DLM: Explicitly Modeling Token Dependencies\\in Diffusion Language Models}

\author{Pengyu Ji\footnotemark[3]\footnotemark[4] \quad Zichen Zhang\footnotemark[3]\footnotemark[4] \quad Xiang Hu\footnotemark[5] \quad Kewei Tu\footnotemark[3]\footnotemark[4]~\thanks{~~Corresponding author.} \\
\footnotemark[3]School of Information Science and Technology, ShanghaiTech University \\
\footnotemark[4]Shanghai Engineering Research Center of Intelligent Vision and Imaging\\
 \tt \{jipy2023, zhangzch2023, tukw\}@shanghaitech.edu.cn \\
\footnotemark[5]Tencent \\
 \tt shawnxxxhu@tencent.com \\}

\begin{document}
\maketitle
\begin{abstract}
Diffusion Language Models (DLMs) generate text by iteratively denoising a masked sequence, independently predicting multiple tokens at each step.
This conditional independence discards inter-token dependencies and degrades coherence---an issue that parallels the multi-modality problem in Non-Autoregressive Translation (NAT).
Drawing on the Directed Acyclic Transformer (DAT), which tackles this problem in NAT via a Directed Acyclic Graph (DAG), we propose DA-DLM, a model that adapts DAG-based dependency modeling to DLMs' iterative setting through a position-oriented DAG design.
The position-oriented DAG binds node groups to fixed output positions so that tokens fixed in earlier steps anchor neighboring predictions via learned transitions, and evolves with denoising to focus on remaining uncertainty as anchors accumulate.
On language modeling, open-ended generation, and summarization, DA-DLM consistently outperforms Block Diffusion, especially under fewer denoising steps, and matches autoregressive models while preserving the parallel generation advantage. Our code is publicly available at \url{https://github.com/jipy0222/DA-DLM}.

\end{abstract}

\input{introduction}
\input{correspondences}
\input{methods}
\input{experiments}
\input{related_works}
\input{conclusion}
\input{limitation}

\section*{Acknowledgments}
This work was supported by the HPC platform of ShanghaiTech University, and
the Core Facility Platform of Computer Science and Communication, SIST, ShanghaiTech University.

\bibliography{custom.bib}
\input{appendix.tex}

\end{document}

%% file: introduction.tex
\section{Introduction}
\label{sec: introduction}

Diffusion Language Models (DLMs)~\cite{nie2025large, bie2025llada2, bie2026llada2} have emerged as a promising paradigm for text generation. 
Most existing DLMs adopt a masked diffusion framework~\cite{sahoo2024simple}, with Block Diffusion (BD3LM)~\cite{arriola2025block} extending it to a block-wise setting. Starting from a fully masked sequence, they progressively recover the target text over multiple denoising steps, each predicting a set of tokens in parallel.
Unlike autoregressive (AR) models that generate tokens sequentially, DLMs are not bound by a fixed generation order and can iteratively refine the sequence with bi-directional context, offering the potential for high-quality and high-speed parallel generation.

However, realizing this potential remains challenging in practice. 
At each denoising step, DLMs only model token-wise marginals, predicting multiple token independently without capturing dependencies among these simultaneously generated tokens.
This conditional independence assumption is a key structural limitation: without coordination, tokens predicted within the same step lack coherence at the sequence level. 
The problem further intensifies when fewer denoising steps are used to improve efficiency, as each step must predict more tokens in parallel \cite{liu2025discrete, wu2025fast, li2026breaking}.

This type of per-position independent prediction method, along with its associated problem, is not unique to DLMs.
In machine translation, Non-Autoregressive Transformers (NATs)~\cite{gu2018nonautoregressive} also independently predict multiple target tokens given a masked or incomplete sequence.
It is well-known that NATs suffer from the multi-modality problem~\cite{gu2018nonautoregressive, huang2022learning}: per-position independent predictions mix fragments from different valid translations, mirroring the coherence issue observed in DLMs.

The connection between DLMs and NATs suggests that techniques for handling the multi-modality problem in the latter can inform solutions to the associated issue in the former.
In this work, we focus on a particular NAT technique, the Directed Acyclic Transformer (DAT)~\cite{huang2022directed}, which tackles this problem by explicitly modeling token dependencies. It generates a Directed Acyclic Graph (DAG) that represents possible tokens and probabilistic transitions between them. By finding a proper path through the DAG, DAT predicts all output tokens in a single step while ensuring they are mutually consistent. 
However, directly transplanting DAT's mechanism into DLMs is non-trivial. In DLMs' iterative denoising, each step resolves a portion of the remaining uncertainty, relying on tokens fixed in earlier steps to effectively anchor subsequent predictions~\cite{sun2025mask, hong2026understanding}, 
yet DAT's dependency structure---a large DAG with unrestricted transitions---is tailored to single-step generation and weakens the ability of previously fixed tokens to guide subsequent predictions. Without effective anchoring, successive denoising steps drift apart rather than converge toward a coherent output.

\begin{figure}
\centering
\includegraphics[width=\linewidth]{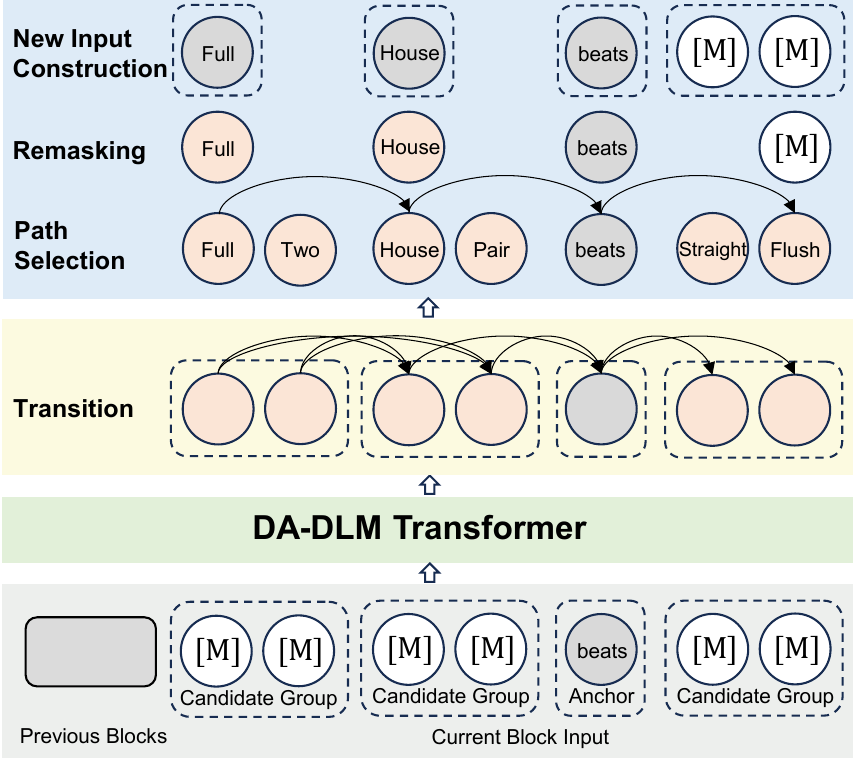}
\caption{Overview of one denoising step in DA-DLM. Given the current block input with one anchor and three candidate groups (bottom), the model produces per-node emissions and inter-group transitions (middle), selects a path, remasks, and yields a new DAG for the next step.}
\label{fig:architecture}
\end{figure}

In this work, we propose the Directed Acyclic Diffusion Language Model (DA-DLM), which adapts the principle of dependency modeling originated from DAT to DLMs' iterative setting.
Built upon BD3LM, DA-DLM constructs a position-oriented DAG at each denoising step, in which every node is tied to a specific output position:
Tokens fixed in earlier steps are pinned at their positions as single-node anchors in the DAG, structurally guiding predictions at surrounding positions. Positions still undetermined are each expanded into a group of candidates tied to it, capturing local modalities. Transitions between adjacent groups encode inter-token dependencies, so that jointly selecting one candidate from each group yields a transition path and hence coordinated predictions rather than independent token choices.
As denoising progresses, the DAG evolves accordingly to reflect the current state of generation: the number of anchors increases while fewer candidate groups remain, gradually refining the graph to resolve the remaining uncertainty.
Figure~\ref{fig:architecture} gives an overview of the model and illustrates how the position-oriented DAG evolves in one denoising step.

We evaluate DA-DLM on language modeling, open-ended generation, and the downstream summarization task.
Results show that DA-DLM consistently outperforms BD3LM baselines across all settings, with particularly large gains when fewer denoising steps are used, and matches the autoregressive baseline on summarization while retaining the parallel decoding advantage.

To summarize, this paper makes the following contributions:
\begin{itemize}[noitemsep]
\vspace{-5pt}
\item We identify a connection between DLMs and NATs, which opens up a promising avenue for drawing inspiration from NAT research to improve DLMs. 
\item Inspired by DAT, a NAT solution designed to address the conditional independence limitation, we propose DA-DLM, which introduces explicit token dependency modeling into DLMs' denoising process, replacing independent per-position prediction with coordinated joint modeling over a position-oriented DAG at each denoising step.
\item We evaluate DA-DLM across diverse tasks and inference scenarios, demonstrating consistent gains over the BD3LM baseline and competitive performance with autoregressive models.
\vspace{-5pt}
\end{itemize}

%% file: correspondences.tex
\section{Background and Motivation}
\label{sec: background}

\subsection{Block Diffusion}
\label{sec: preliminaries}

Block Diffusion (BD3LM)~\cite{arriola2025block} is the backbone of leading DLMs~\cite{bie2025llada2, bie2026llada2}, which we adopt as the foundation for our method.
BD3LM partitions a sequence into $B$ blocks of fixed size $N$ 
and factorizes the likelihood autoregressively across blocks:
\setlength{\abovedisplayskip}{2pt}
\setlength{\belowdisplayskip}{2pt}
\begin{equation}
\log p_\theta(x) = \sum\nolimits_{b=1}^{B} \log p_\theta(x^b \mid x^{<b}).
\label{eq:bd3lm_ar}
\end{equation}

Within each block, $p_\theta(x^b \mid x^{<b})$ is modeled via continuous-time masked diffusion.
The forward noising process $q(\cdot)$ randomly masks tokens in the current block $x^b$ according to the noise schedule $\alpha_t$, producing a partially masked block $x^b_t$ at time $t$.
The preceding blocks $x^{<b}$ serve as conditioning that are not subject to masking.

The denoiser reverses this process under the SUBS parameterization~\cite{sahoo2024simple}: it independently predicts clean tokens at each masked position, while unmasked tokens are carried over unchanged.
The training objective adopts the simplified NELBO of \citet{sahoo2024simple}:
\begin{equation}
\begin{split}
&\mathcal{L}_{BD3LM}(x; \theta) \\
&= \sum_{b=1}^{B} \mathbb{E}_{t, q}\; \frac{\alpha'_t}{1 - \alpha_t} \!\cdot\! \sum_{i} \log p_\theta(x^b_i \mid x^b_t, x^{<b}),
\end{split}
\label{eq:bd3lm_nelbo}
\end{equation}
where $i$ ranges over all block positions, $\alpha'_t = \mathrm{d}\alpha_t / \mathrm{d}t$; only masked positions contribute.

At inference time, each block is generated by iterative denoising from a fully masked sequence: at each step, the model predicts all masked positions in parallel, a subset of predictions is fixed based on a schedule or confidence criterion, and fixed tokens remain unchanged in subsequent steps.

\subsection{Cost of Conditional Independence}
\label{sec: equivalence}

BD3LM's per-position factorized prediction mirrors the approach of Non-Autoregressive Translation (NAT)~\cite{gu2018nonautoregressive}: both independently predict multiple tokens given a masked or incomplete sequence.
The cost of this conditional independence has been formally characterized by \citet{huang2022learning} in the NAT literature.
They show that for any model predicting $M$ tokens independently given arbitrary conditioning $X$:
\begin{equation}
\min\nolimits_\theta D_{\mathrm{KL}}[P_{\mathrm{data}}(Y|X) \| P_\theta(Y|X)] \geq C,
\label{eq:tc_bound}
\end{equation}
where $C = \sum_{i=1}^{M} H(y_i|X) - H(Y|X)$ is the conditional total correlation: the dependency information among target tokens that factorized prediction necessarily discards.
Since BD3LM employs the same factorized prediction per step, this analytical framework applies.
The cost is inherent to per-position independent prediction instead of being specific to either NAT or DLM, manifesting as the multi-modality problem in NAT and as coherence degradation with fewer denoising steps in BD3LM.

\subsection{Dependency Modeling as Missing Piece}
\label{sec: transfer}

\citet{huang2022learning} further extend their analysis to iterative NATs (Appendix~A in \citealp{huang2022learning}), showing that iterative refinement progressively enriches the per-step conditioning, reducing the conditional independence cost (Eq.~\ref{eq:tc_bound}) at each step; both iterative NATs (e.g., CMLM~\cite{ghazvininejad-etal-2019-mask}) and BD3LM already employ this approach.
Their main analysis, on the other hand, focuses on single-step NAT, where the Directed Acyclic Transformer (DAT)~\cite{huang2022directed} stands out by explicitly modeling inter-token dependencies: it constructs a Directed Acyclic Graph (DAG) over the output, where nodes carry token predictions and edges encode probabilistic transitions.
The training objective maximizes the data likelihood marginalized over all paths through the DAG, and inference selects the most probable path, yielding coordinated predictions in a single generation step.
This dependency-based approach offers a complementary path to iterative refinement and is precisely what BD3LM currently lacks.

However, the iterative setting of BD3LM introduces a requirement absent from single-step generation.
In BD3LM, tokens fixed in earlier denoising steps serve as anchors that guide subsequent predictions~\cite{sun2025mask, hong2026understanding}; successive steps build upon previously committed tokens.
DAT's position-free DAG, in which paths may skip arbitrary nodes via unrestricted transitions, dilutes the influence of these anchors: fixed tokens cannot reliably anchor their surrounding predictions, causing successive denoising steps to drift apart rather than converge toward a coherent output.
Adapting dependency modeling to the iterative setting, therefore, requires a structural redesign that respects position stability and the evolving masking state across steps.

%% file: methods.tex
\section{DA-DLM: Directed Acyclic Diffusion Language Model}
\label{sec: method}

As established in \S\ref{sec: transfer}, BD3LM already benefits from iterative refinement but lacks dependency modeling within each step's predictor.
We address this by constructing a position-oriented DAG at each denoising step, whose structure reflects the current masking state and evolves as tokens are progressively determined.

\subsection{Position-Oriented DAG}
\label{sec: dag}

\paragraph{Group-wise structure.}
To incorporate a DAG into BD3LM's per-step predictor, we require a stable correspondence between nodes and output positions that allows fixed tokens to consistently occupy designated nodes across denoising steps.
In a position-free DAG, transitions obey a left-to-right ordering but may skip arbitrary nodes.
Fixed tokens can be enforced as path waypoints---nodes that every valid path must pass through---yet the remaining nodes between waypoints are not bound to fixed output positions and may still realign across denoising steps.

The position-oriented DAG addresses this with a group-wise organization.
For a block of $N$ positions, the DAG consists of $N$ groups, one per position.
Each group $i$ contains $K$ candidate nodes $\{v_i^1, \ldots, v_i^K\}$, each carrying an independent token prediction distribution.
A path $A = (a_1, \ldots, a_N)$, with $a_i \in \{1, \ldots, K\}$, selects exactly one node from each group, producing a sequence of $N$ tokens inherently aligned with the output positions.
The group chain is bookended by a start-of-block and an end-of-block node that provide the initial transition distribution and fixed termination.

\paragraph{Anchors and candidate groups.}
The DAG distinguishes two types of groups reflecting the denoising state.
A position whose token has been fixed in a previous step reduces to a single-node anchor: every path must pass through this node, pinning the known token at its position.
A position still undetermined forms a candidate group with $K$ active nodes, capturing plausible alternatives for that position.
This structure evolves with denoising: as tokens are progressively fixed, candidate groups become anchors, and the DAG focuses on the remaining uncertainty, as illustrated in Figure~\ref{fig:architecture}.

\paragraph{Inter-group transitions.}
Token dependencies are explicitly modeled through transitions between adjacent groups: from each node in group $i$ to the candidates in group $i+1$.
The transition probability is computed via dot-product attention~\cite{huang2022directed}, with softmax normalized over the next group's candidates:
\begin{equation}
\small
T(v_i^j \!\to\! v_{i+1}^k) = \frac{\exp\!\big((\mathbf{q}_i^j)^\top \mathbf{k}_{i+1}^k / \sqrt{d}\big)}{\sum_{k'=1}^{K} \exp\!\big((\mathbf{q}_i^j)^\top \mathbf{k}_{i+1}^{k'} / \sqrt{d}\big)},
\label{eq:transition}
\end{equation}
where $\mathbf{q}_i^j$ and $\mathbf{k}_{i+1}^k$ are query and key projections of the hidden states at nodes $v_i^j$ and $v_{i+1}^k$, respectively.

Since an anchor at position $i$ is the sole node in its group, all incoming transitions from the preceding group converge to it with probability one.
However, its outgoing transitions form a learned distribution over the next group's candidates, allowing the fixed token to directly shape which candidates are favored at the adjacent position in addition to serving as attention context for all positions.

\paragraph{Path probability.}
The joint probability of a target sequence $x^b$ and a path $A$ decomposes into per-node emission probabilities (i.e., token prediction distributions) and inter-group transition probabilities:
\begin{equation}
\begin{split}
&P_\theta(x^b, A \vert x^b_t, x^{<b}) \\
&= \prod_{i=1}^{N}\!p_\theta(x_i^b \vert v_i^{a_i}, x^b_t, x^{<b}) \!\cdot\!\!\! \prod_{i=1}^{N-1}\!\!T(v_i^{a_i} \!\!\to\!\! v_{i+1}^{a_{i+1}}),
\end{split}
\label{eq:path_prob}
\end{equation}

where $x^b_t$ is the partially masked block at time $t$ and $x^{<b}$ denotes the preceding blocks.
Marginalizing over all paths yields the model likelihood $p_\theta(x^b \mid x^b_t, x^{<b}) = \sum_A P_\theta(x^b, A \mid x^b_t, x^{<b})$.

In contrast to DAT's position-free DAG, where a path selects $M$ nodes from $L \gg M$ nodes via a global $L \times L$ transition matrix, DA-DLM restricts each transition to a local $K \times K$ matrix between adjacent groups, and every path has length of exactly $N$---a structural property that maintains position stability across denoising steps.

\subsection{Training}
\label{sec: training}

\paragraph{Training objective.}
DA-DLM retains BD3LM's block-autoregressive factorization (Eq.~\ref{eq:bd3lm_ar}) and replaces the per-position factorized predictor within each block with the DAG-based joint distribution.
Under the same noise-schedule weighting as BD3LM~\cite{sahoo2024simple, arriola2025block}, the training objective becomes:
\begin{equation}
\begin{split}
&\mathcal{L}_{DA-DLM}(x; \theta) \\
&= \sum_{b=1}^{B} \mathbb{E}_{t, q}\!\; \frac{\alpha'_t}{1 - \alpha_t} \!\cdot\! \log \sum_{A} P_\theta(x^b, A \mid x^b_t, x^{<b}).
\end{split}
\label{eq:da_loss}
\end{equation}
Compared to BD3LM's training objective (Eq.~\ref{eq:bd3lm_nelbo}), the per-position independent log-probabilities $\sum_{i} \log p_\theta(x_i^b \mid \cdot)$ are replaced by a single path-marginalized term $\log \sum_A P_\theta(x^b, A \mid \cdot)$.

\paragraph{Efficient marginalization via forward algorithm.}

The objective requires marginalizing over all paths, which is expensive due to the combinatorial number of paths.
We employ a forward algorithm analogous to that in hidden Markov models (details in Appendix~\ref{appendix: algorithm}).
Because transitions only connect adjacent groups, each recursion step aggregates over at most $K$ predecessors within the preceding group, yielding $O(NK^2)$ time complexity.
In contrast, in a position-free DAG~\cite{huang2022directed} with $L{=}NK$ vertices, each node aggregates over all predecessors, resulting in $O(N^3K^2)$ complexity---a factor of $N^2$ higher.

\paragraph{Diffusion training regime.}
Following BD3LM's training protocol~\cite{sahoo2024simple, arriola2025block}, each training step samples a timestep $t$ for each block, and the forward masking process corrupts the target $x^b$ into a partially observed $x^b_t$ by independently masking each token with probability $1 - \alpha_t$.
The resulting masking pattern directly instantiates the position-oriented DAG: unmasked positions form single-node anchors while masked positions expand into candidate groups, and the model is trained to optimize the path-marginalized objective (Eq.~\ref{eq:da_loss}) over this structure.
As $t$ varies across training steps, the model encounters DAGs with different anchor densities---from mostly candidate groups under heavy masking to predominantly anchors under light masking---naturally exposing the model to the full spectrum of partial context.

\subsection{Inference}
\label{sec: inference}

DA-DLM inherits BD3LM's semi-autoregressive generation. 
Within each block, generation proceeds by iterative denoising over $T$ steps, starting from a fully masked sequence.
At each step, the model constructs a DAG from the current masking state, selects a path to obtain token predictions, applies a remasking strategy to determine which tokens become anchors, and optionally refines local predictions via infilling (illustrated in Figure~\ref{fig:architecture}).

\paragraph{Path selection.}
Given emissions and transitions from a forward pass, two decoding strategies operate on the position-oriented DAG.

The Viterbi Strategy finds the jointly optimal path and token sequence~\cite{shao-etal-2022-viterbi}.
Analogous to the training forward algorithm, we apply a max-product recursion over the group chain with the same $O(NK^2)$ complexity (details in Appendix~\ref{appendix: algorithm}).
In a position-free DAG~\cite{shao-etal-2022-viterbi}, the algorithm must additionally search over all possible path lengths and rerank candidates with a tuned length penalty.
Our position-oriented structure eliminates this overhead: the path length is fixed to $N$, requiring no length search or additional hyperparameters, and the time complexity reduces from $O(N^3K^2)$ to $O(NK^2)$.

The Sampling Strategy draws a path from the DAG's distribution (Eq.~\ref{eq:path_prob}) via ancestral sampling: starting from the start-of-block node, we sequentially draw each group's candidate from the current node's outgoing transitions and sample a token from the selected node's emission.

\paragraph{Remasking.}
After predicting tokens along the selected path, a remasking strategy determines which predictions to fix as anchors for the next denoising step.
Standard masked diffusion applies random remasking~\cite{nie2025large, sahoo2024simple, arriola2025block}, where the noise schedule determines the expected number of tokens to unmask at each step and the selection is random; confidence-aware parallel decoding---fixing tokens whose prediction confidence exceeds a threshold---is another widely adopted alternative~\cite{wu2025fast}.
We adapt the confidence-based approach to the position-oriented DAG by defining a composite score that incorporates both the emission and transition information along the selected path:
\begin{equation}
s_i = (1 - w) \cdot p_{\mathrm{emit}}(i) + w \cdot p_{\mathrm{trans}}(i),
\label{eq:confidence}
\end{equation}
where $p_{\mathrm{emit}}(i)$ is the emission probability of the predicted token, $p_{\mathrm{trans}}(i)$ is the average of the incoming and outgoing transition probabilities at position $i$ along the selected path, and $w$ is the adjusted weight.
Positions with $s_i$ above a threshold are fixed as anchors; the rest are remasked.

\paragraph{Infilling.}
As denoising progresses, fixed anchors may partition the remaining masked positions into isolated gaps---contiguous spans bounded by anchors on both sides.
When a gap is sufficiently short, the bounding anchors highly constrain the generation space, so we directly fix predictions within such gaps without remasking.
Infilling is governed by two parameters: \texttt{gap\_length} $k$ and \texttt{context\_ratio} $r$.
It takes effect only when the fraction of already-fixed positions exceeds $r$; at that point, all gaps of length ${\leq}\,k$ are filled in a single step.

%% file: experiments.tex
\section{Experiments}
\label{sec: experiments}

We follow the architecture and training recipe of BD3LM~\cite{arriola2025block}; all models are trained on OpenWebText~\cite{Gokaslan2019OpenWeb} (context length 1024).
DA-DLM uses $K{=}4$ candidates per group; the DAG-specific modules add ${\sim}$2.5M parameters (${\sim}$2\% of the backbone), keeping the model size comparable to BD3LM.
Full training details are in Appendix~\ref{appendix: implementation}.

\subsection{Language Modeling}
\label{subsec: loss}

\paragraph{Evaluation setup.}
Following BD3LM's evaluation protocol~\cite{arriola2025block}, we evaluate on two benchmarks: OpenWebText~\cite{Gokaslan2019OpenWeb} (context length 1024) and One Billion Word (LM1B;~\cite{chelba2013one}, context length 128).
In addition to the OpenWebText models described above, we train a separate set of models on LM1B following the same training procedure.
For each dataset, we compare MDLM and BD3LM with block sizes $\in \{16, 8, 4\}$, and the corresponding DA-DLM variants.
For language modeling evaluation, we use the noise-schedule-weighted average of negative conditional log-likelihoods to measure conditional block reconstruction.
Specifically, the metric averages, over ground-truth blocks $x^b$ and masking patterns, the negative log-likelihood of $x^b$ conditioned on the preceding blocks $x^{<b}$ and its partially masked version $x_t^b$, with weights determined by the noise schedule. 
MDLM and BD3LM compute this negative conditional log-likelihood using token-wise factorization (following Eq.~\ref{eq:bd3lm_nelbo}), whereas DA-DLM computes it by marginalizing over DAG paths (following Eq.~\ref{eq:da_loss}).
Despite the difference in computation, each matched BD3LM/DA-DLM pair uses the same target blocks, conditioning inputs, data, masking process, and noise-schedule weighting. The scores are therefore directly comparable, with lower values indicating better conditional reconstruction quality. 
For MDLM and BD3LM, the score computed using token-wise factorization additionally coincides with the negative evidence lower bound (NELBO), a variational upper bound on the true negative log-likelihood (NLL). This equivalence does not hold for DA-DLM, whose score is computed by marginalizing over DAG paths. We therefore do not convert the reported scores to perplexity or compare them with the AR baseline’s exact NLL.

\begin{table}[tb]
\centering
\small
\setlength{\tabcolsep}{4pt}
\begin{tabular}{@{}llcc@{}}
\hline
\multirow{2}{*}{\textbf{Models}} & \multirow{2}{*}{\textbf{Block Size}} & \multicolumn{2}{c}{\textbf{Conditional Reconstruction ($\downarrow$)}} \\ \cline{3-4}
& & \textbf{LM1B} & \textbf{OpenWebText} \\ \hline\hline
MDLM & --- & 3.47 & 3.15 \\ \hline
BD3LM & 16 & 3.46 & 3.11 \\
& 8 & 3.43 & 3.09 \\
& 4 & 3.39 & 3.04 \\ \hline
DA-DLM & 16 & \textbf{3.30} & \textbf{2.92} \\
& 8 & \textbf{3.28} & \textbf{2.89} \\
& 4 & \textbf{3.24} & \textbf{2.86} \\ \hline\hline
\end{tabular}
\caption{Conditional reconstruction results on language modeling benchmarks. We report the noise-schedule-weighted average of negative conditional log-likelihoods; lower values are better.
}
\label{tab:loss_result}
\end{table}

\paragraph{Results.}
Table~\ref{tab:loss_result} reports language modeling performance on both benchmarks, as measured by the conditional reconstruction metric.
Consistent with \citet{arriola2025block}, smaller block sizes yield better performance; DA-DLM consistently outperforms BD3LM at every block size.
This advantage stems from dependency modeling: whereas BD3LM predicts each position independently, DA-DLM's transition probabilities couple adjacent predictions along paths, concentrating probability mass on coherent token sequences.

\subsection{Open-Ended Generation}
\label{subsec: mauve}

\begin{figure*}
    \centering
    \includegraphics[width=0.9\textwidth]{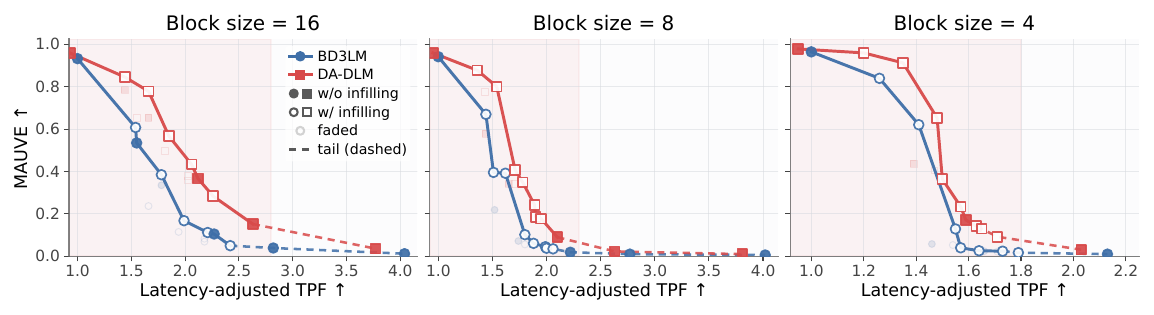}
    \caption{MAUVE--adjusted-TPF trade-off across different block sizes. Each panel reports the Pareto frontier between generation quality (measured by MAUVE) and computational efficiency (measured by adjusted TPF).}
    \label{fig:mauve_tpf_tradeoff}
\end{figure*}

\begin{table}[tb]
\centering
\small
\setlength{\tabcolsep}{0.7pt}
\begin{tabular}{@{}llcccc@{}}
\hline
\multirow{2}{*}{\textbf{Denoising}}
& \multirow{2}{*}{\textbf{Infilling}}
& \multicolumn{2}{c}{\textbf{Adj. TPF} $\uparrow$}
& \multicolumn{2}{c}{\textbf{MAUVE: $\mathbf{B=16}$} $\uparrow$} \\ \cline{3-4}\cline{5-6}
& & \textbf{BD3LM} & \textbf{DA-DLM}
& \textbf{BD3LM} & \textbf{DA-DLM} \\ \hline\hline
FH & --         & 1.00 & 0.93 & 0.933 & \textbf{0.960} \\
FH & $(2, 0.5)$ & 1.54 & 1.44 & 0.608 & \textbf{0.846} \\
FH & $(2, 0.0)$ & 1.78 & 1.66 & 0.385 & \textbf{0.779} \\
FH & $(3, 0.5)$ & 1.66 & 1.55 & 0.237 & \textbf{0.652} \\ \hline
$T=16$ & --         & 1.55 & 1.44 & 0.535 & \textbf{0.786} \\
$T=16$ & $(2, 0.5)$ & 1.99 & 1.85 & 0.168 & \textbf{0.567} \\
$T=16$ & $(2, 0.0)$ & 1.94 & 1.81 & 0.115 & \textbf{0.495} \\
$T=16$ & $(3, 0.5)$ & 2.18 & 2.03 & 0.083 & \textbf{0.377} \\ \hline
$T=12$ & --         & 1.78 & 1.66 & 0.335 & \textbf{0.653} \\
$T=12$ & $(2, 0.5)$ & 2.21 & 2.06 & 0.112 & \textbf{0.434} \\
$T=12$ & $(2, 0.0)$ & 2.18 & 2.03 & 0.069 & \textbf{0.358} \\
$T=12$ & $(3, 0.5)$ & 2.42 & 2.26 & 0.050 & \textbf{0.284} \\ \hline
$T=8$ & -- & 2.27 & 2.12 & 0.105 & \textbf{0.367} \\
$T=6$ & -- & 2.82 & 2.63 & 0.039 & \textbf{0.152} \\
$T=4$ & -- & 4.04 & 3.77 & 0.012 & \textbf{0.036} \\ \hline\hline
\end{tabular}
\caption{MAUVE--adjusted TPF comparison between BD3LM and DA-DLM on OpenWebText generation with block size $\mathbf{B}=16$. The BD3LM column reports raw TPF, while the DA-DLM column reports latency-adjusted TPF (Adj. TPF) calibrated using the measured per-step latency ratio (\S\ref{subsec: inference}). The MAUVE score of the AR model is $0.981$, for reference.}
\label{tab:mauve_tpf_bs16}
\end{table}

\paragraph{Evaluation setup.}
We evaluate open-ended generation quality with prefix-conditioned generation on OpenWebText and measure the MAUVE score~\cite{pillutla2021mauve} (details in Appendix~\ref{appendix: generation_setup}).
We compare an AR baseline, BD3LM with block sizes $\in \{16, 8, 4\}$, and the corresponding DA-DLM variants.
To map the quality--efficiency trade-off, we vary the number of denoising steps $T$ per block and compute raw tokens per forward pass (TPF), defined as the average number of tokens determined per forward pass at runtime. 
For the quality--efficiency comparison, we report latency-adjusted TPF (Adj. TPF). At each block size, DA-DLM's raw TPF is calibrated using the ratio of BD3LM's per-step latency to that of DA-DLM, with BD3LM serving as the latency reference. The latency measurements are reported in \S\ref{subsec: inference}.
We evaluate two sampling mechanisms: (1) first-hitting sampling (FH)~\cite{zheng2025masked}, where each forward pass determines one token (raw TPF$=1$); and (2) random remasking~\cite{sahoo2024simple, nie2025large}, where the noise schedule determines the expected number of tokens to unmask at each step.
DA-DLM additionally employs the sampling path selection strategy (\S\ref{sec: inference}) to decode from the DAG at each denoising step.
We also evaluate both models with and without the infilling mechanism (\S\ref{sec: inference}), parameterized by gap length $k$ and context ratio $r$ (reported as $(k, r)$ in Table~\ref{tab:mauve_tpf_bs16}).

\paragraph{Results.}
Table~\ref{tab:mauve_tpf_bs16} reports detailed results for block size 16. 
Under first-hitting sampling (raw TPF${=}1$), both models achieve comparable MAUVE scores. As decoding becomes more parallel, yielding higher adjusted TPF, BD3LM's generation quality drops sharply, while DA-DLM maintains substantially higher MAUVE. 
When the number of denoising steps becomes very small, both models eventually degrade.

Infilling further differentiates the two models: for BD3LM, infilling largely follows the existing quality--efficiency tradeoff curve, offering little advantage beyond what reducing denoising steps already provides.
For DA-DLM, infilling yields a favorable tradeoff when gap regions are well-constrained by surrounding anchors---for instance, when combined with sufficient denoising steps or when the context ratio gates infilling on adequate surrounding context.
DA-DLM's transition probabilities propagate dependency information from anchors into gaps, making infilling an effective accelerator that expands DA-DLM's Pareto frontier to higher adjusted TPF values.

Figure~\ref{fig:mauve_tpf_tradeoff} extends this analysis across all three block sizes (detailed numerical results in Tables~\ref{tab:mauve_tpf_bs16_detail}--\ref{tab:mauve_tpf_bs4_detail}, Appendix~\ref{appendix: results}), where each panel plots the Pareto frontier between MAUVE and adjusted TPF. Solid lines trace the Pareto frontier, while dashed segments mark the high-efficiency tail (both models eventually degrade when the number of denoising steps becomes very small); filled markers denote configurations without infilling, and open markers denote infilling variants. 
The shaded region highlights DA-DLM's operating regime---the range of adjusted TPF over which it maintains a favorable quality--efficiency trade-off. 
The trends observed for block size 16 hold consistently: DA-DLM's Pareto frontier surpasses BD3LM's across all block sizes, and infilling configurations extend DA-DLM's frontier further into regions with higher adjusted TPF. Faded markers indicate configurations dominated within each model's own frontier.
Smaller block sizes achieve higher absolute MAUVE under first-hitting sampling, but degrade more steeply under random remasking due to cross-block error compounding: smaller blocks partition the sequence into more blocks, so each block's parallel generation errors accumulate more rapidly through the conditioning prefix, creating an increasing distribution shift from the ground-truth prefixes seen in training.
DA-DLM mitigates this by maintaining higher per-block quality, reducing the error that propagates into downstream blocks.

\begin{table}[tb]
\centering
\small
\setlength{\tabcolsep}{3pt}
\begin{tabular}{@{}llcc@{}}
\hline
\multirow{2}{*}{\textbf{Denoising}}
& \multirow{2}{*}{\textbf{Infilling}}
& \multicolumn{2}{c}{\textbf{LLM-as-a-Judge: $\mathbf{B=16}$}} \\ \cline{3-4}
& & \textbf{DA-DLM Win (\%)} $\uparrow$
& \textbf{95\% CI (\%)} \\ \hline\hline

FH & --         & 52.39 & [50.31, 54.47] \\
FH & $(2, 0.5)$ & 63.56 & [61.48, 65.59] \\
FH & $(2, 0.0)$ & 69.40 & [67.45, 71.29] \\
FH & $(3, 0.5)$ & 71.06 & [69.04, 72.99] \\ \hline

$T=16$ & --         & 65.38 & [63.33, 67.37] \\
$T=16$ & $(2, 0.5)$ & 73.48 & [71.53, 75.35] \\
$T=16$ & $(2, 0.0)$ & 76.47 & [74.58, 78.26] \\
$T=16$ & $(3, 0.5)$ & 74.94 & [72.98, 76.80] \\ \hline

$T=12$ & --         & 68.85 & [66.83, 70.81] \\
$T=12$ & $(2, 0.5)$ & 72.45 & [70.45, 74.36] \\
$T=12$ & $(2, 0.0)$ & 76.96 & [75.07, 78.75] \\
$T=12$ & $(3, 0.5)$ & 75.59 & [73.59, 77.47] \\ \hline

$T=8$ & -- & 72.07 & [70.01, 74.04] \\
$T=6$ & -- & 73.31 & [71.20, 75.32] \\
$T=4$ & -- & 74.02 & [71.66, 76.25] \\ \hline\hline
\end{tabular}
\caption{Pairwise LLM-as-a-judge evaluation of DA-DLM against BD3LM on OpenWebText generation with block size $\mathbf{B}=16$. DA-DLM Win denotes the percentage of DA-DLM wins among non-tied pairs. Confidence intervals are 95\% Wilson intervals.}
\label{tab:llm_judge_bs16}
\end{table}

\begin{table*}[tb]
\centering
\small
\setlength{\tabcolsep}{4pt}
\begin{tabular}{llcccccccccc}
\hline
\multirow{3}{*}{\textbf{Models}} & \multirow{3}{*}{\textbf{Strategy}} & \multicolumn{10}{c}{\textbf{Summarization}} \\ \cline{3-12}
& & \multicolumn{5}{c}{\textbf{XSum}} & \multicolumn{5}{c}{\textbf{CNN/DailyMail}} \\ \cline{3-12}
& & \textbf{R-1} & \textbf{R-2} & \textbf{R-L} & \textbf{Avg} & \textbf{TPF} & \textbf{R-1} & \textbf{R-2} & \textbf{R-L} & \textbf{Avg} & \textbf{TPF} \\ \hline\hline
AR & Greedy & 35.72 & \textbf{13.51} & \textbf{28.31} & 25.85 & \multicolumn{1}{c|}{1.00} & 38.54 & 16.36 & \textbf{26.57} & 27.16 & 1.00 \\ \hline
\multirow{2}{*}{BD3LM} & Conf 0.9 & 36.05 & 13.26 & 27.64 & 25.65 & \multicolumn{1}{c|}{1.07} & 38.21 & \textbf{16.85} & 25.95 & 27.00 & 1.81 \\
 & Conf 0.9 + Infill & 36.10 & 13.17 & 27.67 & 25.65 & \multicolumn{1}{c|}{1.20} & 38.21 & 16.83 & 25.94 & 26.99 & 1.92 \\ \hline
\multirow{2}{*}{DA-DLM} & Conf 0.9 & 36.44 & 13.15 & 28.06 & 25.88 & \multicolumn{1}{c|}{1.13} & \textbf{38.60} & 16.76 & 26.16 & \textbf{27.17} & 1.73 \\
 & Conf 0.9 + Infill & \textbf{36.55} & 13.14 & 28.09 & \textbf{25.93} & \multicolumn{1}{c|}{\textbf{1.50}} & 38.53 & 16.72 & 26.10 & 27.12 & \textbf{2.56} \\ \hline\hline
\end{tabular}
\caption{Results on summarization tasks. `R-1/2/L' denotes ROUGE-1 / ROUGE-2 / ROUGE-L scores. 
`Conf 0.9' means confidence-aware parallel decoding with threshold 0.9, and `Infill' denotes infilling enabled.}
\label{tab:summarization_result}
\end{table*}

\paragraph{LLM-as-a-Judge Evaluation.}
To complement MAUVE, we additionally assess the quality and coherence of generated continuations through an LLM-as-a-Judge evaluation. 
\verb|DeepSeek-V4-Flash| serves as the pairwise judge for the generations evaluated in \S\ref{subsec: mauve}. For each decoding configuration, the prompt and corresponding anonymized continuations from DA-DLM and BD3LM are presented in both A/B orders to control for position bias. Under conservative aggregation, a win is recorded only when both orders favor the same model, and we report DA-DLM’s win rate among non-tied pairs. Further details of the judging protocol and aggregation rule are provided in Appendix~\ref{appendix: llm_judge_setup}.
Table~\ref{tab:llm_judge_bs16} reports the pairwise evaluation results for block size 16, while Tables~\ref{tab:llm_judge_bs16_appendix}--\ref{tab:llm_judge_bs4_appendix} in Appendix~\ref{appendix: results} present the complete results across all three block sizes $\in \{16, 8, 4\}$.
The judge consistently favors DA-DLM over BD3LM under matched decoding configurations across all block sizes, with the preference generally becoming stronger in more aggressively parallel decoding regimes and when infilling is enabled. This trend broadly aligns with MAUVE and provides complementary evidence that explicit dependency modeling improves generation quality and coherence.

\subsection{Downstream Task: Summarization}
\label{subsec: summary}

\paragraph{Evaluation setup.}
We evaluate on two summarization datasets---XSum~\cite{narayan-etal-2018-dont} and CNN/DailyMail~\cite{nallapati-etal-2016-abstractive}. 
Models pre-trained on OpenWebText are fine-tuned for this task.
Based on the observation from \S\ref{subsec: mauve} that models with smaller block sizes struggle to maintain generation quality over long sequences, we select a block size of 16 for both BD3LM and DA-DLM.
We report ROUGE~\cite{lin-hovy-2003-automatic} and raw tokens per forward pass (TPF) as quality and efficiency metrics; decoding configurations are detailed in Appendix~\ref{appendix: summarization_setup}.

\paragraph{Results.}
Table~\ref{tab:summarization_result} reports ROUGE scores and raw TPF on both summarization benchmarks.
DA-DLM outperforms BD3LM in average ROUGE on both datasets and 
remains competitive with the AR baseline, demonstrating that the benefits of dependency modeling transfer to task-specific generation after fine-tuning.
The advantage concentrates on ROUGE-1 and ROUGE-L, suggesting that DA-DLM's joint modeling over paths helps the model stay on topic and maintain sequential coherence.

Both BD3LM and DA-DLM generate multiple tokens per forward pass under confidence-aware decoding alone, indicating strong task adaptation. Infilling consistently increases the degree of parallelism on both datasets with minimal impact on quality: average ROUGE changes by less than 0.05 in either direction.

\subsection{Inference Efficiency}
\label{subsec: inference}

\begin{table}[tb]
\centering
\small
\setlength{\tabcolsep}{4pt}
\begin{tabular}{@{}lccccc@{}}
\hline
\multirow{2}{*}{\textbf{Model}}
& \multirow{2}{*}{$\mathbf{B}$}
& \multirow{2}{*}{\textbf{Latency (ms)} $\downarrow$}
& \multicolumn{3}{c}{\textbf{Memory (MiB)}} \\ \cline{4-6}
& & & \textbf{Start} & \textbf{End} & \textbf{Peak} $\downarrow$ \\ \hline\hline
BD3LM
& 16 & 27.59 & 1306 & 2440 & 3814 \\
& 8  & 22.49 & 1306 & 2449 & 3136 \\
& 4  & 20.51 & 1306 & 2457 & 2987 \\ \hline
DA-DLM
& 16 & 29.60 & 1325 & 2459 & 3890 \\
& 8  & 23.73 & 1324 & 2467 & 3183 \\
& 4  & 21.48 & 1324 & 2475 & 3019 \\ \hline\hline
\end{tabular}
\caption{Per-step latency and allocated GPU memory for BD3LM and
DA-DLM. Here $\mathbf{B}$ denotes the block size; Start and End denote memory
allocation before prompt prefill and after generation, respectively,
while Peak denotes the maximum allocation during generation.}
\label{tab:inference_efficiency}
\end{table}

\paragraph{Setup.}
We profile BD3LM and DA-DLM on a single NVIDIA A800 GPU, following the open-ended generation setting in \S\ref{subsec: mauve} with a batch size of 32. We evaluate block sizes $\in \{16, 8, 4\}$. Infilling is disabled; both models use first-hitting sampling, while DA-DLM additionally applies its sampling-based path selection. Per-step latency measures the complete denoising update, including model computation and the corresponding sampling and remasking operations. We average latency over 10 measured runs after 10 warm-up runs, with CUDA synchronization around each timed update. GPU memory is profiled separately, and we report the allocated memory before prompt prefill, after generation, and at its peak. 

\paragraph{Results.}
As shown in Table~\ref{tab:inference_efficiency}, DA-DLM introduces modest per-step latency overheads of 2.01, 1.24, and 0.97 ms over BD3LM for block sizes 16, 8, and 4, respectively, corresponding to relative increases of 7.3\%, 5.5\%, and 4.7\%. Its additional peak memory usage is at most 76 MiB, or 2.0\%. The additional latency arises from candidate-transition modeling and path sampling in DA-DLM, but remains small relative to the denoising computation. These results show that the two models have similar per-step inference costs under matched decoding settings.

The measured per-step latency ratios are used to calibrate raw TPF and obtain the adjusted TPF values reported in \S\ref{subsec: mauve}. For example, at block size 16, DA-DLM's raw TPF is scaled by $27.59/29.60=0.932$. After applying the corresponding calibration at each block size, DA-DLM retains its quality--efficiency advantage in the MAUVE comparisons reported in \S\ref{subsec: mauve}.

%% file: related_works.tex
\section{Related Work}
\label{sec: related work}

\subsection{Directed Acyclic Transformers}
\label{subsec: rw_dat}

Since the introduction of DAT~\cite{huang2022directed}, a number of follow-up works have improved its theoretical understanding~\cite{huang2022learning}, decoding~\cite{shao-etal-2022-viterbi}, architectures~\cite{li-etal-2024-non,li2025dat}, and training objectives~\cite{an2023optimizing, ma2023fuzzy}.

PreDAT~\cite{huang-etal-2023-directed} extends DAT from machine translation to diverse text generation tasks via pre-training, sharing our goal of broadening DAG-based generation beyond translation, yet it remains within the encoder-decoder paradigm.
Diff-DAT~\cite{nguyen-tri-etal-2025-diffusion}, focusing on seq2seq machine translation, shares with our work the idea of extending DAT from single-step to iterative generation via a diffusion process.
However, both PreDAT and Diff-DAT retain DAT's position-free DAG, which dilutes the anchoring effect of fixed tokens across denoising steps.
Consequently, PreDAT remains a single-step generator, and Diff-DAT reports that iterative decoding degrades beyond two steps.
DA-DLM addresses this limitation with the position-oriented DAG and further departs from the seq2seq paradigm by building on block diffusion~\cite{arriola2025block} in a decoder-only architecture for general-purpose language modeling.

\subsection{Dependency Modeling in DLMs}
\label{subsec: rw_dep}

Recent work on the conditional independence limitation in DLMs follows two lines.

The first retains the factorized predictor and supplements it with an external or static dependency source:
DCD~\cite{liu2025discrete} couples the diffusion model's marginals with an external pretrained autoregressive model at inference time, and CoDD~\cite{li2026breaking} augments the output distribution with a static Probabilistic Circuit once trained.
Neither condition the dependency information on the evolving denoising state.

The second operates on the decoding side: DEMASK~\cite{ringel2026dependency} and DAPD~\cite{kim2026dapd} estimate pairwise token dependencies and selectively unmask only approximately independent subsets at each step.

DA-DLM differs from both: its DAG-based transitions are produced by the same backbone conditioned on the current denoising state, making dependency modeling dynamic; and rather than avoiding co-prediction of dependent tokens, it explicitly models their joint distribution through path marginalization.

%% file: conclusion.tex
\section{Conclusion}
\label{sec: conclusion}

We presented DA-DLM, which introduces explicit inter-token dependency modeling into diffusion language models by constructing a position-oriented DAG at each denoising step.
By bridging the connection between DLMs and NATs, we adapted the DAG-based joint prediction mechanism from DAT to the iterative denoising setting.
The key design is the position-oriented DAG, which binds each node group to a fixed output position so that tokens fixed in earlier steps serve as anchors that reliably guide surrounding predictions---resolving the incompatibility of DAT's position-free DAG with iterative denoising.
Experiments on language modeling, open-ended generation, and summarization show that DA-DLM consistently outperforms the BD3LM baseline, with particularly pronounced gains when fewer denoising steps are used, and achieves competitive performance with autoregressive models while retaining the parallel decoding advantage.

%% file: limitation.tex
\section*{Limitations}

Although DA-DLM can be obtained by up-training from an existing BD3LM checkpoint without training from scratch, our experiments are conducted at a relatively small model scale due to computational constraints.
The evaluations focus on the traditional DLM benchmarks, conditional log-likelihood, open-ended generation quality, and a single downstream task, but do not yet extend to the broader set of capabilities expected of modern language models, such as mathematical reasoning or code generation.
As larger open-source discrete diffusion models become available~\cite{nie2025large, bie2025llada2, bie2026llada2}, up-training on these checkpoints and evaluating on a wider range of benchmarks would provide a more comprehensive validation of the approach.

%% file: appendix.tex
\appendix

\section{Algorithm Details}
\label{appendix: algorithm}

\paragraph{Forward algorithm.}
We recurrently calculate the probability sum of path prefixes that end at candidate $k$ in group $i$ while generating the first $i$ target tokens, denoted as $f_{i,k}$.
Since transitions only connect adjacent groups, a path prefix ending at candidate $k$ in group $i$ must come from some candidate $j$ in the preceding group $i{-}1$, so $f_{i,k}$ can be obtained from $\{f_{i-1,j}\}_{j=1}^{K}$.
By advancing one group at a time, we obtain the probability sum of all valid paths as $\sum_{k=1}^{K} f_{N,k}$.

\paragraph{Viterbi decoding.}
Analogous to the forward algorithm, we recurrently calculate the maximum joint score of path prefixes ending at candidate $k$ in group $i$, denoted $g_{i,k}$.
Since transitions only connect adjacent groups, $g_{i,k}$ is obtained by taking the best predecessor from group $i{-}1$, with the score incorporating both the transition to $v_i^k$ and the best token prediction at that node.
After processing all $N$ groups, we backtrack from the highest-scoring candidate in the final group to recover the optimal path, and select tokens by taking argmax at each node along the path.

\section{General Setup}
\label{appendix: implementation}

All models---including the MDLM, BD3LM, and autoregressive (AR) baselines---share a Transformer backbone with 12 layers, 768 hidden dimensions, and 12 attention heads, following the architecture of BD3LM~\cite{arriola2025block}.
Data is tokenized with the GPT-2 tokenizer~\cite{radford2019language} (vocabulary size 50{,}257).
The AR baseline is trained for half the number of steps to match the total number of tokens seen~\cite{sahoo2024simple}.

For diffusion models, training proceeds in two stages: an initial MDLM~\cite{sahoo2024simple} phase (850k steps, full-sequence diffusion without block structure) followed by a block-wise up-training phase (150k steps) under varying block sizes.
For DA-DLM, the block-wise phase consists of 90k steps of BD3LM training followed by 60k steps of DA-DLM training that introduces the DAG modules (\S\ref{sec: method}), totaling the same 150k up-training budget.

\section{Open-Ended Generation Setup}
\label{appendix: generation_setup}

We randomly select validation sequences from OpenWebText, condition on the first 64 tokens as a prompt, and generate the remaining 960 tokens, yielding completions of length 1024.
We generate 5{,}000 samples per configuration and compute the MAUVE score against the reference completions, using GPT-2 Large~\cite{radford2019language} as the embedding model.
All models employ nucleus sampling with $p{=}0.9$.
The number of denoising steps per block is varied as follows: $\{16, 12, 8, 6, 4\}$ for block size 16; $\{8, 6, 4, 3, 2\}$ for block size 8; $\{4, 3, 2\}$ for block size 4.
All diffusion models in our experiments are time-agnostic, i.e., they do not condition on the diffusion timestep~\cite{sahoo2024simple, ou2025your}.
This allows caching the denoising network's output when no new tokens are unmasked between consecutive steps, so the effective number of forward passes can be smaller than the nominal number of denoising steps. Raw TPF reflects this reduction in the effective number of forward passes, while the adjusted TPF reported in \S\ref{subsec: mauve} additionally incorporates the measured per-step latency ratios.
For the infilling mechanism, we test gap lengths $k \in \{2, 3\}$ with and without a context ratio $r{=}0.5$.

\section{LLM-as-a-Judge Evaluation Setup}
\label{appendix: llm_judge_setup}

We use \verb|DeepSeek-V4-Flash| with temperature 0 and thinking disabled as the pairwise judge. The evaluation uses the OpenWebText prompts and continuations described in \S\ref{subsec: mauve}. For each pair, the model identities are hidden, and the DA-DLM and BD3LM continuations are presented as continuations A and B. Each pair is evaluated in both A/B orders to control for position bias. The judge selects A, B, or a tie based on continuity with the prompt, coherence and logical flow, fluency and readability, specificity, and the absence of generation artifacts. The prompt and each continuation are capped at 1,200 and 6,000 characters, respectively. We use conservative aggregation across the two presentation orders. A win is assigned to a model only when both orders favor that model; all other outcomes are treated as ties. We report DA-DLM's win rate among non-tied pairs, together with 95\% Wilson confidence intervals.

\section{Summarization Setup}
\label{appendix: summarization_setup}

Documents and summaries are truncated to 768 and 256 tokens, respectively, and diffusion models are fine-tuned with a batch size of 512 for 5k steps; the AR model is fine-tuned for half the number of steps, consistent with the pre-training configuration.
Because summarization emphasizes generation accuracy over diversity, we adopt deterministic or high-confidence decoding strategies.
The AR model uses greedy decoding.
For BD3LM, we adopt confidence-aware parallel decoding~\cite{wu2025fast} with a threshold of 0.9.
For DA-DLM, we combine Viterbi path selection with confidence-aware parallel decoding, where the composite confidence score (Eq.~\ref{eq:confidence}) with $w{=}0.5$ integrates transition information into the remasking decision.
We additionally evaluate both diffusion models with the infilling mechanism (\S\ref{sec: inference}, $k{=}2$, $r{=}0.5$).

\section{Detailed Experimental Results}
\label{appendix: results}

\paragraph{Training computation alignment.}
Although BD3LM and DA-DLM share the same 150k-step block-wise up-training budget, DA-DLM's per-step FLOPs are higher due to the expanded input from the DAG's candidate nodes.
To rule out the possibility that DA-DLM's gains merely reflect greater total training compute, we compare against BD3LM-300K, a BD3LM variant trained for 300k block-wise up-training steps to match DA-DLM's total training FLOPs.
Table~\ref{tab:mauve_tpf_bs16_300k} shows that DA-DLM still consistently outperforms BD3LM-300K across all configurations, confirming that the improvements stem from the DAG-based dependency modeling rather than from additional training compute.

\clearpage

\begin{table*}[tb]
\centering
\setlength{\tabcolsep}{6pt}
\begin{tabular}{llcccc}
\hline
\multirow{2}{*}{\textbf{Denoising}}
& \multirow{2}{*}{\textbf{Infilling}}
& \multicolumn{2}{c}{\textbf{Adj. TPF} $\uparrow$}
& \multicolumn{2}{c}{\textbf{MAUVE: Block Size = 16} $\uparrow$} \\ \cline{3-4}\cline{5-6}
& & \textbf{BD3LM} & \textbf{DA-DLM}
& \textbf{BD3LM} & \textbf{DA-DLM} \\ \hline\hline
FH & -- & 1.00 & 0.93 & 0.933 & \textbf{0.960} \\
FH & $k=2, r=0.5$ & 1.54 & 1.44 & 0.608 & \textbf{0.846} \\
FH & $k=2$ & 1.78 & 1.66 & 0.385 & \textbf{0.779} \\
FH & $k=3, r=0.5$ & 1.66 & 1.55 & 0.237 & \textbf{0.652} \\ \hline
$T=16$ & -- & 1.55 & 1.44 & 0.535 & \textbf{0.786} \\
$T=16$ & $k=2, r=0.5$ & 1.99 & 1.85 & 0.168 & \textbf{0.567} \\
$T=16$ & $k=2$ & 1.94 & 1.81 & 0.115 & \textbf{0.495} \\
$T=16$ & $k=3, r=0.5$ & 2.18 & 2.03 & 0.083 & \textbf{0.377} \\ \hline
$T=12$ & -- & 1.78 & 1.66 & 0.335 & \textbf{0.653} \\
$T=12$ & $k=2, r=0.5$ & 2.21 & 2.06 & 0.112 & \textbf{0.434} \\
$T=12$ & $k=2$ & 2.18 & 2.03 & 0.069 & \textbf{0.358} \\
$T=12$ & $k=3, r=0.5$ & 2.42 & 2.26 & 0.050 & \textbf{0.284} \\ \hline
$T=8$ & -- & 2.27 & 2.12 & 0.105 & \textbf{0.367} \\
$T=6$ & -- & 2.82 & 2.63 & 0.039 & \textbf{0.152} \\
$T=4$ & -- & 4.04 & 3.77 & 0.012 & \textbf{0.036} \\ \hline\hline
\end{tabular}
\caption{MAUVE--adjusted TPF comparison between BD3LM and DA-DLM on OpenWebText generation with block size 16. The BD3LM column reports raw TPF, while the DA-DLM column reports latency-adjusted TPF (Adj. TPF) calibrated using the measured per-step latency ratio (\S\ref{subsec: inference}). Boldface denotes the higher MAUVE score under the same decoding strategy. Here $T$ denotes the number of denoising steps, $k$ denotes the infill gap length, and $r$ denotes the context ratio. MAUVE for the AR model is $0.981$, for reference.}
\label{tab:mauve_tpf_bs16_detail}
\end{table*}

\begin{table*}[tb]
\centering
\setlength{\tabcolsep}{6pt}
\begin{tabular}{llcccc}
\hline
\multirow{2}{*}{\textbf{Denoising}}
& \multirow{2}{*}{\textbf{Infilling}}
& \multicolumn{2}{c}{\textbf{Adj. TPF} $\uparrow$}
& \multicolumn{2}{c}{\textbf{MAUVE: Block Size = 8} $\uparrow$} \\ \cline{3-4}\cline{5-6}
& & \textbf{BD3LM} & \textbf{DA-DLM}
& \textbf{BD3LM} & \textbf{DA-DLM} \\ \hline\hline
FH & -- & 1.00 & 0.95 & 0.942 & \textbf{0.960} \\
FH & $k=2, r=0.5$ & 1.44 & 1.36 & 0.670 & \textbf{0.878} \\
FH & $k=2$ & 1.62 & 1.54 & 0.392 & \textbf{0.801} \\
FH & $k=3, r=0.5$ & 1.51 & 1.43 & 0.396 & \textbf{0.776} \\ \hline
$T=8$ & -- & 1.52 & 1.44 & 0.219 & \textbf{0.578} \\
$T=8$ & $k=2, r=0.5$ & 1.80 & 1.71 & 0.102 & \textbf{0.407} \\
$T=8$ & $k=2$ & 1.80 & 1.71 & 0.056 & \textbf{0.343} \\
$T=8$ & $k=3, r=0.5$ & 1.88 & 1.78 & 0.061 & \textbf{0.349} \\ \hline
$T=6$ & -- & 1.74 & 1.65 & 0.072 & \textbf{0.341} \\
$T=6$ & $k=2, r=0.5$ & 1.99 & 1.89 & 0.045 & \textbf{0.242} \\
$T=6$ & $k=2$ & 2.00 & 1.90 & 0.038 & \textbf{0.185} \\
$T=6$ & $k=3, r=0.5$ & 2.06 & 1.95 & 0.035 & \textbf{0.178} \\ \hline
$T=4$ & -- & 2.22 & 2.10 & 0.019 & \textbf{0.090} \\
$T=3$ & -- & 2.77 & 2.63 & 0.010 & \textbf{0.023} \\
$T=2$ & -- & 4.02 & 3.81 & 0.006 & \textbf{0.010} \\ \hline\hline
\end{tabular}
\caption{MAUVE--adjusted TPF comparison between BD3LM and DA-DLM on OpenWebText generation with block size 8. The BD3LM column reports raw TPF, while the DA-DLM column reports latency-adjusted TPF (Adj. TPF) calibrated using the measured per-step latency ratio (\S\ref{subsec: inference}). Boldface denotes the higher MAUVE score under the same decoding strategy. Here $T$ denotes the number of denoising steps, $k$ denotes the infill gap length, and $r$ denotes the context ratio. MAUVE for the AR model is $0.981$, for reference.}
\label{tab:mauve_tpf_bs8_detail}
\end{table*}

\begin{table*}[tb]
\centering
\setlength{\tabcolsep}{6pt}
\begin{tabular}{llcccc}
\hline
\multirow{2}{*}{\textbf{Denoising}}
& \multirow{2}{*}{\textbf{Infilling}}
& \multicolumn{2}{c}{\textbf{Adj. TPF} $\uparrow$}
& \multicolumn{2}{c}{\textbf{MAUVE: Block Size = 4} $\uparrow$} \\ \cline{3-4}\cline{5-6}
& & \textbf{BD3LM} & \textbf{DA-DLM}
& \textbf{BD3LM} & \textbf{DA-DLM} \\ \hline\hline
FH & -- & 1.00 & 0.95 & 0.965 & \textbf{0.979} \\
FH & $k=2, r=0.5$ & 1.26 & 1.20 & 0.840 & \textbf{0.960} \\
FH & $k=2$ & 1.41 & 1.35 & 0.621 & \textbf{0.913} \\
FH & $k=3$ & 1.55 & 1.48 & 0.129 & \textbf{0.653} \\ \hline
$T=4$ & -- & 1.46 & 1.39 & 0.058 & \textbf{0.436} \\
$T=4$ & $k=2, r=0.5$ & 1.54 & 1.47 & 0.053 & \textbf{0.405} \\
$T=4$ & $k=2$ & 1.57 & 1.50 & 0.040 & \textbf{0.364} \\
$T=4$ & $k=3$ & 1.64 & 1.57 & 0.026 & \textbf{0.234} \\ \hline
$T=3$ & -- & 1.66 & 1.59 & 0.024 & \textbf{0.171} \\
$T=3$ & $k=2, r=0.5$ & 1.71 & 1.63 & 0.022 & \textbf{0.142} \\
$T=3$ & $k=2$ & 1.73 & 1.65 & 0.024 & \textbf{0.130} \\
$T=3$ & $k=3$ & 1.79 & 1.71 & 0.017 & \textbf{0.092} \\ \hline
$T=2$ & -- & 2.13 & 2.03 & 0.010 & \textbf{0.031} \\ \hline\hline
\end{tabular}
\caption{MAUVE--adjusted TPF comparison between BD3LM and DA-DLM on OpenWebText generation with block size 4. The BD3LM column reports raw TPF, while the DA-DLM column reports latency-adjusted TPF (Adj. TPF) calibrated using the measured per-step latency ratio (\S\ref{subsec: inference}). Boldface denotes the higher MAUVE score under the same decoding strategy. Here $T$ denotes the number of denoising steps, $k$ denotes the infill gap length, and $r$ denotes the context ratio. MAUVE for the AR model is $0.981$, for reference.}
\label{tab:mauve_tpf_bs4_detail}
\end{table*}

\begin{table*}[tb]
\centering
\setlength{\tabcolsep}{6pt}
\begin{tabular}{llcccc}
\hline
\multirow{2}{*}{\textbf{Denoising}}
& \multirow{2}{*}{\textbf{Infilling}}
& \multicolumn{2}{c}{\textbf{Adj. TPF} $\uparrow$}
& \multicolumn{2}{c}{\textbf{MAUVE: Block Size = 16} $\uparrow$} \\ \cline{3-4}\cline{5-6}
& & \textbf{BD3LM-300K} & \textbf{DA-DLM}
& \textbf{BD3LM-300K} & \textbf{DA-DLM} \\ \hline\hline
FH & -- & 1.00 & 0.93 & 0.927 & \textbf{0.960} \\
FH & $k=2, r=0.5$ & 1.54 & 1.44 & 0.558 & \textbf{0.846} \\
FH & $k=2$ & 1.78 & 1.66 & 0.382 & \textbf{0.779} \\
FH & $k=3, r=0.5$ & 1.66 & 1.55 & 0.264 & \textbf{0.652} \\ \hline
$T=16$ & -- & 1.55 & 1.44 & 0.511 & \textbf{0.786} \\
$T=16$ & $k=2, r=0.5$ & 1.99 & 1.85 & 0.184 & \textbf{0.567} \\
$T=16$ & $k=2$ & 1.94 & 1.81 & 0.107 & \textbf{0.495} \\
$T=16$ & $k=3, r=0.5$ & 2.18 & 2.03 & 0.098 & \textbf{0.377} \\ \hline
$T=12$ & -- & 1.78 & 1.66 & 0.325 & \textbf{0.653} \\
$T=12$ & $k=2, r=0.5$ & 2.21 & 2.06 & 0.124 & \textbf{0.434} \\
$T=12$ & $k=2$ & 2.18 & 2.03 & 0.076 & \textbf{0.358} \\
$T=12$ & $k=3, r=0.5$ & 2.42 & 2.26 & 0.058 & \textbf{0.284} \\ \hline
$T=8$ & -- & 2.27 & 2.12 & 0.091 & \textbf{0.367} \\
$T=6$ & -- & 2.82 & 2.63 & 0.034 & \textbf{0.152} \\
$T=4$ & -- & 4.04 & 3.77 & 0.011 & \textbf{0.036} \\ \hline\hline
\end{tabular}
\caption{MAUVE--adjusted TPF comparison between BD3LM-300K and DA-DLM on OpenWebText generation with block size 16. The BD3LM column reports raw TPF, while the DA-DLM column reports latency-adjusted TPF (Adj. TPF) calibrated using the measured per-step latency ratio (\S\ref{subsec: inference}). Boldface denotes the higher MAUVE score under the same decoding strategy. Here $T$ denotes the number of denoising steps, $k$ denotes the infill gap length, and $r$ denotes the context ratio. MAUVE for the AR model is $0.981$, for reference.}
\label{tab:mauve_tpf_bs16_300k}
\end{table*}

\begin{table*}[tb]
\centering
\setlength{\tabcolsep}{6pt}
\begin{tabular}{llcc}
\hline
\multirow{2}{*}{\textbf{Denoising}}
& \multirow{2}{*}{\textbf{Infilling}}
& \multicolumn{2}{c}{\textbf{LLM-as-a-Judge: Block Size = 16}} \\ \cline{3-4}
& & \textbf{DA-DLM Win Rate (\%)} $\uparrow$
& \textbf{95\% CI (\%)} \\ \hline\hline
FH & --                 & 52.39 & [50.31, 54.47] \\
FH & $k=2, r=0.5$       & 63.56 & [61.48, 65.59] \\
FH & $k=2$              & 69.40 & [67.45, 71.29] \\
FH & $k=3, r=0.5$       & 71.06 & [69.04, 72.99] \\ \hline
$T=16$ & --             & 65.38 & [63.33, 67.37] \\
$T=16$ & $k=2, r=0.5$   & 73.48 & [71.53, 75.35] \\
$T=16$ & $k=2$          & 76.47 & [74.58, 78.26] \\
$T=16$ & $k=3, r=0.5$   & 74.94 & [72.98, 76.80] \\ \hline
$T=12$ & --             & 68.85 & [66.83, 70.81] \\
$T=12$ & $k=2, r=0.5$   & 72.45 & [70.45, 74.36] \\
$T=12$ & $k=2$          & 76.96 & [75.07, 78.75] \\
$T=12$ & $k=3, r=0.5$   & 75.59 & [73.59, 77.47] \\ \hline
$T=8$  & --             & 72.07 & [70.01, 74.04] \\
$T=6$  & --             & 73.31 & [71.20, 75.32] \\
$T=4$  & --             & 74.02 & [71.66, 76.25] \\ \hline\hline
\end{tabular}
\caption{LLM-as-a-judge comparison between DA-DLM and BD3LM on
OpenWebText generation with block size 16. Here $T$ denotes the
number of denoising steps, $k$ denotes the infill gap length, and
$r$ denotes the context ratio. DA-DLM win rates are computed among
non-tied pairs, and the reported confidence intervals (CIs) are
95\% Wilson intervals.}
\label{tab:llm_judge_bs16_appendix}
\end{table*}

\begin{table*}[tb]
\centering
\setlength{\tabcolsep}{6pt}
\begin{tabular}{llcc}
\hline
\multirow{2}{*}{\textbf{Denoising}}
& \multirow{2}{*}{\textbf{Infilling}}
& \multicolumn{2}{c}{\textbf{LLM-as-a-Judge: Block Size = 8}} \\ \cline{3-4}
& & \textbf{DA-DLM Win Rate (\%)} $\uparrow$
& \textbf{95\% CI (\%)} \\ \hline\hline
FH & --                 & 53.54 & [51.56, 55.52] \\
FH & $k=2, r=0.5$       & 66.36 & [64.41, 68.26] \\
FH & $k=2$              & 69.55 & [67.61, 71.43] \\
FH & $k=3, r=0.5$       & 69.74 & [67.80, 71.61] \\ \hline
$T=8$ & --              & 71.70 & [69.70, 73.62] \\
$T=8$ & $k=2, r=0.5$    & 76.10 & [74.18, 77.92] \\
$T=8$ & $k=2$           & 78.40 & [76.54, 80.15] \\
$T=8$ & $k=3, r=0.5$    & 78.94 & [77.07, 80.70] \\ \hline
$T=6$ & --              & 76.33 & [74.38, 78.18] \\
$T=6$ & $k=2, r=0.5$    & 77.28 & [75.32, 79.13] \\
$T=6$ & $k=2$           & 81.32 & [79.48, 83.04] \\
$T=6$ & $k=3, r=0.5$    & 78.92 & [76.97, 80.73] \\ \hline
$T=4$ & --              & 76.68 & [74.58, 78.66] \\
$T=3$ & --              & 77.21 & [74.83, 79.42] \\
$T=2$ & --              & 72.06 & [68.92, 75.00] \\ \hline\hline
\end{tabular}
\caption{LLM-as-a-judge comparison between DA-DLM and BD3LM on
OpenWebText generation with block size 8. Here $T$ denotes the
number of denoising steps, $k$ denotes the infill gap length, and
$r$ denotes the context ratio. DA-DLM win rates are computed among
non-tied pairs, and the reported confidence intervals (CIs) are
95\% Wilson intervals.}
\label{tab:llm_judge_bs8_appendix}
\end{table*}

\begin{table*}[tb]
\centering
\setlength{\tabcolsep}{6pt}
\begin{tabular}{llcc}
\hline
\multirow{2}{*}{\textbf{Denoising}}
& \multirow{2}{*}{\textbf{Infilling}}
& \multicolumn{2}{c}{\textbf{LLM-as-a-Judge: Block Size = 4}} \\ \cline{3-4}
& & \textbf{DA-DLM Win Rate (\%)} $\uparrow$
& \textbf{95\% CI (\%)} \\ \hline\hline
FH & --                 & 55.57 & [53.55, 57.57] \\
FH & $k=2, r=0.5$       & 66.91 & [64.98, 68.77] \\
FH & $k=2$              & 70.33 & [68.44, 72.17] \\
FH & $k=3$              & 80.73 & [79.08, 82.28] \\ \hline
$T=4$ & --              & 84.02 & [82.42, 85.49] \\
$T=4$ & $k=2, r=0.5$    & 86.52 & [85.02, 87.89] \\
$T=4$ & $k=2$           & 88.77 & [87.38, 90.03] \\
$T=4$ & $k=3$           & 88.63 & [87.21, 89.91] \\ \hline
$T=3$ & --              & 86.91 & [85.34, 88.33] \\
$T=3$ & $k=2, r=0.5$    & 86.49 & [84.90, 87.93] \\
$T=3$ & $k=2$           & 87.27 & [85.70, 88.68] \\
$T=3$ & $k=3$           & 88.40 & [86.90, 89.76] \\ \hline
$T=2$ & --              & 87.45 & [85.63, 89.06] \\ \hline\hline
\end{tabular}
\caption{LLM-as-a-judge comparison between DA-DLM and BD3LM on
OpenWebText generation with block size 4. Here $T$ denotes the
number of denoising steps, $k$ denotes the infill gap length, and
$r$ denotes the context ratio. DA-DLM win rates are computed among
non-tied pairs, and the reported confidence intervals (CIs) are
95\% Wilson intervals.}
\label{tab:llm_judge_bs4_appendix}
\end{table*}